\documentclass[preprint,12pt,sort&compress]{elsarticle}

\usepackage{textcomp,booktabs}
\usepackage[usenames,dvipsnames]{color}
\usepackage{colortbl}
\definecolor{mygray}{gray}{.9}
\definecolor{mypink}{rgb}{.99,.91,.95}
\definecolor{mycyan}{cmyk}{.3,0,0,0}
\usepackage{amssymb}
\usepackage{graphicx}
\usepackage{booktabs}
\usepackage{tabularx}
\usepackage{array}
\usepackage{lscape}
\usepackage{longtable}
\usepackage{multirow}
\usepackage{amsmath}
\usepackage{colortbl}
\usepackage[figuresright]{rotating}
\usepackage{epsfig}
\usepackage{epsf}
\usepackage{subfigure}
\usepackage{threeparttable}

\newcommand{\PreserveBackslash}[1]{\let\temp=\\#1\let\\=\temp}
\newcolumntype{C}[1]{>{\PreserveBackslash\centering}p{#1}}
\newcolumntype{R}[1]{>{\PreserveBackslash\raggedleft}p{#1}}
\newcolumntype{L}[1]{>{\PreserveBackslash\raggedright}p{#1}}
\newtheorem{definition}{Definition}[section]

\journal{} 
\begin{document}

\begin{frontmatter}



\title{A quantum dynamic belief decision making model}


\author[address1]{Zichang He}
\author[address1]{Wen Jiang\corref{label1}}
\address[address1]{School of Electronics and Information, Northwestern Polytechnical University, Xi'an, Shaanxi, 710072, China}
\cortext[label1]{Corresponding author at Wen Jiang: School of Electronics and Information, Northwestern Polytechnical University, Xi'an, Shaanxi 710072, China. Tel: (86-29)88431267. E-mail address: jiangwen@nwpu.edu.cn, jiangwenpaper@hotmail.com}

\begin{abstract}
The sure thing principle and the law of total probability are basic laws in classic probability theory. A disjunction fallacy leads to the violation of these two classical probability laws. In this paper, a new quantum dynamic belief decision making model based on quantum dynamic modelling and Dempster-Shafer (D-S) evidence theory is proposed to address this issue and model the real human decision-making process. Some mathematical techniques are borrowed from quantum mathematics. Generally, belief and action are two parts in a decision making process. The uncertainty in belief part is represented by a superposition of certain states. The uncertainty in actions is represented as an extra uncertainty state. The interference effect is produced due to the entanglement between beliefs and actions. Basic probability assignment (BPA) of decisions is generated by quantum dynamic modelling. Then BPA of the extra uncertain state and an entanglement degree defined by an entropy function named Deng entropy are used to measure the interference effect. Compared the existing model, the number of free parameters is less in our model. Finally, a classical categorization decision-making experiment is illustrated to show the effectiveness of our model.
\end{abstract}

\end{frontmatter}
\section{Introduction}\label{Introduction}
The sure thing principle introduced by Jim Savage\cite{Jarrett1956The} is a fundamental principle in economics and probability theory. It means that if one prefers action $A$ over $B$ under state of the world $X$, while action $A$ is also preferred under the opposite state of world $\neg X$, then it can be concluded that one will still prefer action $A$ over $B$ under the state of the world is unspecified. The law of total probability is a fundamental rule relating marginal probabilities to conditional probabilities. It expresses the total probability of an outcome which can be realized via several distinct events. However, many experiments and studies have shown that the sure thing principle and the law of total probability can be violated due to the disjunction effect. The disjunction fallacy is an empirical finding in which the proportion taking the target gamble under the unknown condition falls below both of the proportions taking the target gamble under each of the known conditions. The same person takes the target gamble under both known conditions, but then rejects the target gamble under the unknown condition\cite{Lambdin2007The}.

Generally, the sure thing principle and the law of total probability are basic probability laws. However, the disjunction fallacy leads to the violation of them. To explain it, many studies have been proposed. The original explanation was a psychological idea based on the failure of consequential reasoning under unknown condition\cite{Tversky1992The}. More recently, to explain these paradoxical findings, the theory of quantum probability has been introduced in the quantum cognition and decision making process. Quantum information has a wide application, like in user security\cite{yu2015enhancing}, quantum communication\cite{song2015finite} and so on. Quantum probability is an effective approach to psychology\cite{wang2013potential,pothos2015structured,aerts2013concepts,blutner2013quantum}. It has been widely applied to psychology and decision making fields by explaining order effect\cite{trueblood2011quantum,wang2013quantum,wang2014context}, disjunction effect\cite{pothos2009quantum}, the interference effect of categorization\cite{wang2016interference}, prisoner's dilemma\cite{chen2003quantum}, conceptual combinations\cite{bruza2015probabilistic,aerts2009quantum}, quantum game theory\cite{Situ2016Relativistic,Situ2016Two,Asano2010Quantum} and so on. To explain the disjunction fallacy which leads to the violation of the sure thing principle and the law of total probability, many models have been proposed, such as a quantum dynamical model\cite{Pothos2009A,Busemeyer2009Empirical}, quantum prospect decision theory\cite{Yukalov2015Quantum,Yukalov2009Physics,Yukalov2009Processing,Yukalov2011Decision} and quantum-like Bayesian networks\cite{Moreira2016Quantum} etc. Besides quantum-like approach toolbox was also proposed\cite{Denolf2016Bohr,Nyman2011On,Nyman2011Quantum}.

In this paper, a new quantum dynamic belief decision making model based on quantum dynamic modelling and D-S evidence theory is proposed to explain the disjunction fallacy. Dempster-Shafer evidence theory was proposed by Dempster in 1967\cite{Dempster1967Upper} and modified by Shafer in 1978\cite{Shafer1978A}. Evidential reasoning is an approach handing the evidence is D-S theory\cite{Yang1994An,Fu2014Determining}. And Deng entropy\cite{dengentropy} is an efficient tool to measure the information volume of evidence\cite{jiang2016AN,Fei2016Meausre}. Many applications in realistic projects have shown the power of D-S evidence theory handling uncertain information\cite{Ma2015An,Fu2015A,Du2014New,Su2015Combining}. Besides, it has been applied in quantum information and quantum probability in many works\cite{Vourdas2014Lower,Vourdas2014Quantum,He2012Classification,Resconi2001Tests}. In our model, D-S theory is used to extend the action state space and uncertainty during the decision-making process is the crucial factor of measuring the interference effect.

Consider the prisoner dilemma (PD) paradigm\cite{Tversky1992The}, two players need to decide independently whether to cooperate with opponent or to defect against opponent. Beliefs and actions are two parts in a decision-making process. The real situation in action part is that some participants can not make a precise choice to cooperate or to defect for sure, but they are forced to make a final decision in the experiment. In our model, the uncertainty in action is represented in a specific uncertain state and the uncertainty in belief is represented by a superposition of certain states. As the beliefs and actions are entangled in some degree, the interference effect can be produced. Unlike the previous models, the interference effect is measured by the distribution of the uncertain state in action. The quantum dynamic model is applied to generate the basic probability assignment (BPA), which measures the supporting of a decision. Then the uncertain information will be distributed by using Deng entropy. A classical categorization decision-making experiment is illustrated in this paper to show the effectiveness of our model. Though an extra uncertain state is introduced, the new model is more succinct as the free parameters decrease compared with classical quantum dynamic models. Because the entanglement degree is calculated by an entropy function rather being set as a free parameter.

The rest of the paper is organized as follows. In Section 2, the preliminaries of basic theories employed are briefly introduced. The new quantum dynamic belief decision making model is proposed in Section 3. Then a categorization decision-making experiment is illustrated and our new model is applied to it in Section 4. Finally, Section 5 comes to the conclusion.
\section{Preliminaries}
\subsection{Quantum dynamic model}
The quantum dynamic model model first proposed by Busemeyer $et al.$ in 2006\cite{Busemeyer2006Quantum} is formulated as a random walk decision process. The evolution of complex valued probability amplitudes over time is described. The interference effect can be generated in a quantum model which is not possible in a classical Markov model.
The quantum dynamic model assumes that a participant has some potential to be in every state in the beginning. Thus the person's state is a superposition of all possible $n$ states
\[\left| \psi  \right\rangle  = {\psi _1}\left| {{S_1}} \right\rangle  + {\psi _2}\left| {{S_2}} \right\rangle  +  \ldots {\psi _n}\left| {{S_n}} \right\rangle \]
and the initial state corresponds to an amplitude distribution $\psi \left( {\rm{0}} \right)$ represented by the $n \times 1$ matrix.
\[\psi \left( {\rm{0}} \right){\rm{ = }}\left[ {\begin{array}{*{20}{c}}
{{\psi _{\rm{1}}}}\\
{{\psi _2}}\\
 \vdots \\
{{\psi _n}}
\end{array}} \right].\]
During the decision making process, the state will evolve across time obeying a Schr${\ddot o}$dinger equation.
\begin{equation}\label{schrodinger}
\frac{d}{{dt}}\psi \left( t \right) =  - i \cdot H \cdot \psi \left( t \right)
\end{equation}
where $H$ is a Hamiltonian matrix: ${H^\dag } = H$, $H$ has elements ${h_ij}$ in row $i$ column $j$ representing the instantaneous rate of change to $\left| i \right\rangle $ from $\left| j \right\rangle $.
Eq. (\ref{schrodinger}) has a matrix exponential solution:
\begin{equation}\label{psit_2}
\psi \left( {{t_2}} \right) = {e^{ - iHt}} \cdot \psi \left( {{t_1}} \right){\rm{ = }}U \cdot \psi \left( {{t_1}} \right)
\end{equation}
where matrix $U = {e^{ - iHt}}$ is a unitary matrix: ${U^\dag }U = I$. It finally guarantees that $\psi \left( t \right)$ always has unit length.

For $t = {t_2} - {t_1}$, the transition probabilities ${T_{ij}}$ of observing state $i$ at time ${t_2}$ given that state $j$ was observed at time ${t_1}$ is determined as
\begin{equation}
{T_{ij}}\left( t \right) = {\left| {{U_{ij}}\left( t \right)} \right|^2}
\end{equation}
where ${U_{ij}}$ is the line $i$, column $j$ element of the unitary matrix $U$.

Based on the above definition, the amplitude distribution of person's state evolves to $\psi \left( t \right)$ from the initial $\psi \left( 0 \right)$ across time $t$ as Eq. (\ref{psit_new}), which shows the dynamic in a decision making process.
\begin{equation}\label{psit_new}
\psi \left( t \right) = U \cdot \psi \left( 0 \right)
\end{equation}
\subsection{Dempster-Shafer evidence theory}
Let $F$ denote a finite set composed of all possible values of the random variable $X$. The elements of set $F$ are mutually exclusive. $F$ is called the frame of discernment. Let ${2^F}$ denote the power set of $F$ whose each element corresponds to a subset of values of $X$. Basic probability assignment (BPA) is a mapping from ${2^F}$ to $\left[ {0,1} \right]$, defined as\cite{Dempster1967Upper,Shafer1978A}:
\begin{equation}
m:{2^F} \to \left[ {0,1} \right]
\end{equation}
satisfying
\begin{equation}
\sum\limits_{A \in {2^F}} {m\left( A \right)}  = 1
\end{equation}
and
\begin{equation}
m\left( \emptyset  \right) = 0{\kern 1pt} {\kern 1pt}
\end{equation}
The mass function $m$ represents supporting degree to proposition $A$. A mass function corresponds to a belief ($Bel$) function and a plausibility($Pl$) function respectively.

Given $m:{2^U} \to \left[ {0,1} \right]$, $Bel\left( A \right)$ function represents the whole belief degree to the proposition $A$, defined as
\begin{equation}
Bel\left( A \right) = \sum\limits_{B \subseteq A} {m\left( B \right)} {\kern 1pt} {\kern 1pt} {\kern 1pt} {\kern 1pt} {\kern 1pt} {\kern 1pt} {\kern 1pt} {\kern 1pt} {\kern 1pt} {\kern 1pt} {\kern 1pt} {\kern 1pt} {\kern 1pt} {\kern 1pt} {\kern 1pt} {\kern 1pt} {\kern 1pt} {\kern 1pt} {\kern 1pt} \forall A \subseteq {2^F}
\end{equation}
$Pl$ function represents the belief degree of not denying proposition $A$, defined as
\begin{equation}
Pl\left( A \right) = 1 - Bel\left( {\bar A } \right) = \sum\limits_{B \cap A \ne \emptyset } {m\left( B \right)} {\kern 1pt} {\kern 1pt} {\kern 1pt} {\kern 1pt} {\kern 1pt} {\kern 1pt} {\kern 1pt} {\kern 1pt} {\kern 1pt} {\kern 1pt} {\kern 1pt} {\kern 1pt} {\kern 1pt} {\kern 1pt} {\kern 1pt} {\kern 1pt} {\kern 1pt} {\kern 1pt} \forall A \subseteq {2^F}
\end{equation}
where $\bar A = {2^F}  - A$.

As $Pl\left( A \right) \ge Bel\left( A \right)$, $Pl$ function and $Bel$ function are upper and lower limit of the supporting of $A$.

\begin{figure}[!ht]
\centering
\includegraphics[scale=0.65]{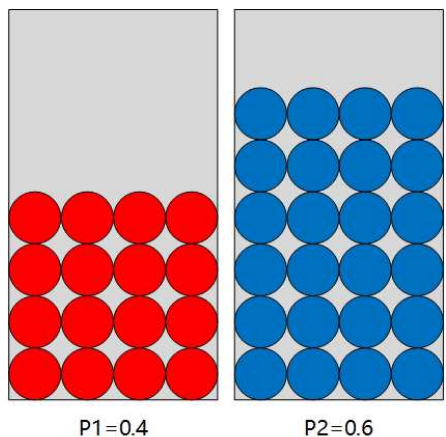}
\caption{A game of picking ball which can be handled by probability theory}\label{randb}
\end{figure}

In the following, an game of picking ball will be used to show the D-S theory's ability of handling uncertainty. There are two boxes filled with some balls as shown in Fig. \ref{randb}. Left box is contended with red balls and right box is contended with blue balls. The number of balls in each box is unknown. Now, a ball is picked randomly from two boxes. The probability of picking from left box $P1$ is known as 0.4 while picking from right box $P2$ is known as 0.6. It is easy to obtain that the probability of picking a red ball is 0.4 while picking a blue ball is 0.6 based on probability theory.

\begin{figure}[!ht]
\centering
\includegraphics[scale=0.65]{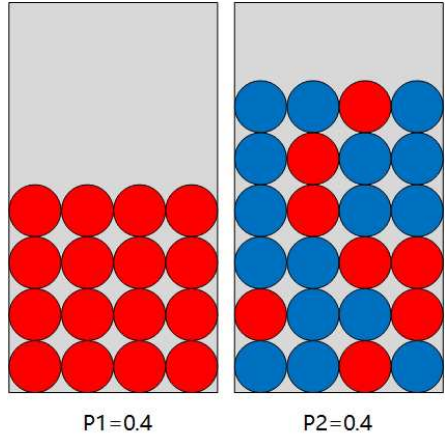}
\caption{A game of picking ball where probability theory is unable but D-S evidence theory is able to handle}\label{randrb}
\end{figure}

Now, the situation changes as shown in Fig. \ref{randrb}. The left box is contended with right balls while the right box is contended with red and blue balls. The exact number of the balls in each box is still unknown and the ratio of red balls with blue balls is completely unknown. The probabilities of selecting from two boxes keep the same,
$P1 = 0.4$ and $P2 = 0.6$. The question is what the probability that a red ball is picked is. Due to the lack of information, the question can not be addressed in probability theory. However, D-S evidence theory can effectively handle it. We can obtain a BPA that $m\left( R \right) = 0.4$ and $m\left( {R,B} \right) = 0.6$. Then the uncertainty is well expressed in the frame of D-S theory.

\subsection{Pignistic probability transformation}
The term "pignistic" proposed by Smets\cite{Smets1994The} is originated from the word pignus, meaning bet in Latin. Pignistic probability transformation (PPT) has a wide application in decision making. Let $m$ be a BPA on a frame of discernment $F$, then the PPT function is defined as
\begin{equation}{
Bet\left( {A} \right) = \sum\nolimits_{A \subseteq B} {\frac{{m\left( {B} \right)}}{{\left| {B} \right|}}}
}\end{equation}
where ${\left| {B} \right|}$ denotes the number of elements in set $B$. This is called as the cardinality of $B$.
\subsection{Deng entropy}
In order to measure the information volume of a BPA, Deng entropy has been proposed\cite{dengentropy}, which is defined as follows
\begin{equation}\label{dengentropy}
{E_d} =  - \sum\limits_i {m\left( {{X_i}} \right){{\log }_2}\frac{{m\left( {{X_i}} \right)}}{{{2^{\left| {{X_i}} \right|}} - 1}}}
\end{equation}
where ${{X_i}}$ is a proposition in BPA $m$ and ${\left| {{X_i}} \right|}$ is the cardinality of ${{X_i}}$ representing the number of elements in it.
For example, for BPA ${m_1}: {m_1}\left( a \right) = {m_1}\left( b \right) = {m_1}\left( {ab} \right) = \frac{1}{3}$, where proposition $ab$ means that both proposition $a$ and $b$ are possible, namely it is uncertain whether belongs to $a$ or $b$.
\[{E_d}\left( {{m_1}} \right) =  - \frac{1}{3} \times {\log _2}{\textstyle{{{1 \mathord{\left/
 {\vphantom {1 3}} \right.
 \kern-\nulldelimiterspace} 3}} \over {{2^1} - 1}}} - \frac{1}{3} \times {\log _2}{\textstyle{{{1 \mathord{\left/
 {\vphantom {1 3}} \right.
 \kern-\nulldelimiterspace} 3}} \over {{2^1} - 1}}} - \frac{1}{3} \times {\log _2}{\textstyle{{{1 \mathord{\left/
 {\vphantom {1 3}} \right.
 \kern-\nulldelimiterspace} 3}} \over {{2^2} - 1}}} = 2.11\]
For BPA ${m_2}: {m_2}\left( a \right) = {m_2}\left( b \right) = \frac{1}{2}$
\[{E_d}\left( {{m_2}} \right) =  - \frac{1}{2} \times {\log _2}{\textstyle{{{1 \mathord{\left/
 {\vphantom {1 2}} \right.
 \kern-\nulldelimiterspace} 2}} \over {{2^1} - 1}}} - \frac{1}{2} \times {\log _2}{\textstyle{{{1 \mathord{\left/
 {\vphantom {1 2}} \right.
 \kern-\nulldelimiterspace} 2}} \over {{2^1} - 1}}} = 1\]

Specially, when the BPA is consisted of singleton sets as ${m_2}$, namely $\left| {{X_i}} \right|$ of all the proposition in BPA is 1. Deng entropy degenerates to Shannon entropy.
\begin{equation}\label{Shannon entropy}
{E_d} =  - \sum\limits_i {m\left( {{X_i}} \right){{\log }_2}\frac{{m\left( {{X_i}} \right)}}{{{2^1} - 1}}}  =  - \sum\limits_i {m\left( {{X_i}} \right){{\log }_2}} m\left( {{X_i}} \right)
\end{equation}
\subsection{Quantum entanglement in D-S evidence theory}
In this part, the correlation between Deng entropy and entanglement is discussed.

Let us consider an example as follows. Suppose 32 students participated a course examination and one of them won the first place. In order to know who is the first one, we go to ask their course teacher. But the teacher does not want to directly tell us. Instead, she just answers “Yes” or “No” to our questions. The problem is how many times do we need ask at most in order to know who is the first one? Assume the times is $t$, it is easy to answer the problem through calculating the information volume by using information entropy
\[t = {\log _2}32 = 5\]
Now, let’s consider another situation. Assume we have been told that there are students tied for first. In this case, how many times do we need ask
at most to know who are the first ones?
In this case, obviously
\[t \ge {\log _2}32\]

According to Deng entropy, the information volume is as follows
\[\begin{array}{l}
{E_d} =  - \frac{1}{{{2^{32}} - 1}} \times {\log _2}\left( {\frac{{{\textstyle{1 \over {{2^{32}} - 1}}}}}{{{2^1} - 1}}} \right) - \frac{1}{{{2^{32}} - 1}} \times {\log _2}\left( {\frac{{{\textstyle{1 \over {{2^{32}} - 1}}}}}{{{2^2} - 1}}} \right)\\
{\kern 1pt} {\kern 1pt} {\kern 1pt} {\kern 1pt} {\kern 1pt} {\kern 1pt} {\kern 1pt} {\kern 1pt} {\kern 1pt} {\kern 1pt} {\kern 1pt} {\kern 1pt} {\kern 1pt} {\kern 1pt} {\kern 1pt} {\kern 1pt}  -  \cdots  - \frac{1}{{{2^{32}} - 1}} \times {\log _2}\left( {\frac{{{\textstyle{1 \over {{2^{32}} - 1}}}}}{{{2^{32}} - 1}}} \right) \approx 48
\end{array}\]

In Deng and Deng (2014)\cite{Deng2014On}, the conclusion is that we need 32 times to determine the top 1 student(s). However, it is not correct since that, according to the result above, we need 48 times to obtain the result. Deng and Deng proposed that entanglement is the key to cause the difference of these two values. Because the entanglement brings a larger information volume and the extra information is measured by Deng entropy.

\section{Proposed method}\label{Proposed method}
\begin{figure}[!ht]
\centering
\includegraphics[scale=0.65]{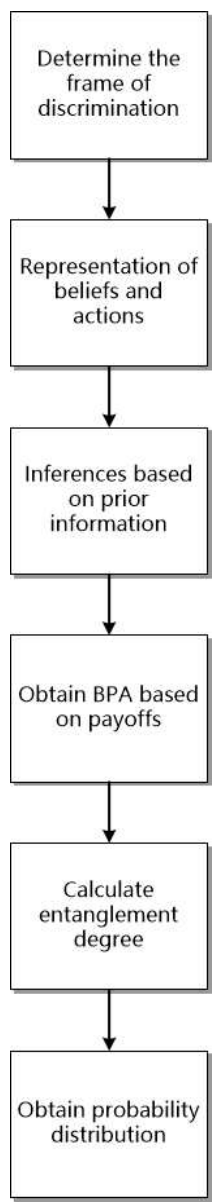}
\caption{The flow chart of proposed model}\label{flowchart}
\end{figure}

The quantum dynamic belief decision making model based on quantum dynamic modelling and D-S evidence theory can model the decision making process. Beliefs and actions are two main parts in a decision making process.
The uncertainty of beliefs is represented by a superposition of certain states which collapses when the beliefs become certain. When the beliefs are uncertain, the interference effect is produced due to the entanglement between beliefs and actions. The concept of interference effect in quantum cognition is that the states can interference with each other, which leads to the violation of the law of total probability.
The uncertainty of actions is expressed as an extra uncertain state with D-S theory. It is rational to assume that some people hesitate to make a precise action decision during a decision making process. However, a final action decision has to be made as an uncertain action is not an outcome. In classical theory, uncertain information in actions can be distributed into basic probabilities using PPT\cite{Smets1994The}. However, in some cases the beliefs and actions are entangled in some degree, which means that beliefs and actions need to be coordinated. It has been verified that participants feel the need to coordinate beliefs and actions.\cite{Krueger2012Social,busemeyer2012Social}. In a entangled system, uncertainty in actions should be handled differently to measure the interference effect. To address it, an entanglement degree is defined in our model using Deng entropy. The flow chart of our model is shown in Fig. \ref{flowchart}.

In the following, an example of PD paradigm is used to illustrate the proposed model. In a PD game, participants need to decide whether to cooperate (C) or to defect (D) against the opponent independently. In the study, participants are requested to make a decision under the condition that the opponent's decision is known and that the opponent's decision is totally unknown. The result is that the probability of defecting under known condition is larger than the one under unknown condition, which violates the law of total probability. Based on the flow chart, the proposed model can be listed step by step as follows:

\textbf{Step 1: Determine the frame of discrimination}

When the opponent's decision is known to defect (D) or to cooperate (C), the outcome of the game will be either to defect (D) or to cooperate (C). Thus the basic states "DD", "CD", "DC" and "DD" consist the frame of discrimination, where, for example, "CD" represents the participant decide to cooperate knowing the opponent defects. Then extra state "CDD", "CDC" and $\emptyset$ will be included to fill the power set of this discrimination. State "CDD" represents that the participant is uncertain to cooperate or defect when knowing the opponent defects and it can be denoted as "UD". Same as it, state "CDC" can be denoted as "UC". And the empty set $\emptyset$ is irrespective in the model as it is meaningless.

When the opponent's decision is totally unknown, the outcome of the game will be either to defect (D) or to cooperate (C). Thus the basic states "DU" and "CU" consist the frame of discrimination, where, for example, "DU" represents the participant decide to defect unknowing the opponent's decision. Then extra state "CDU" and $\emptyset$ will be included to fill the power set of this discrimination. State "CDU" presents that the participant is uncertain to cooperate or defect when unknowing the opponent's decision and it can be denoted as "UU". And the empty set $\emptyset$ is irrespective in the model as it is meaningless.

\textbf{Step 2: Representation of beliefs and actions}

\begin{figure}
\centering
\includegraphics[scale=0.5]{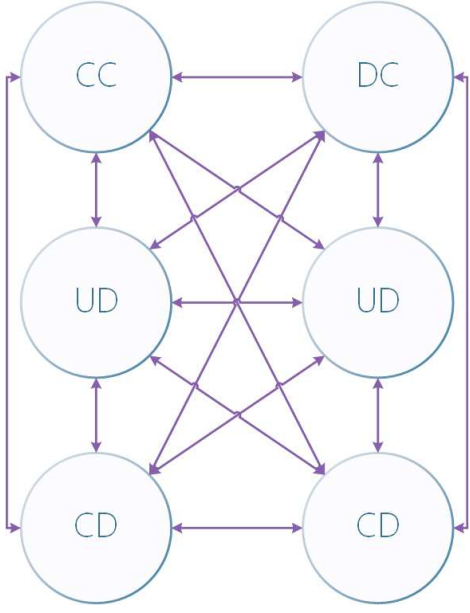}
\caption{Transition diagram in the PD game}\label{sixstate}
\end{figure}

As the states have been extended in Step 1, the initial state involves six combination of beliefs and actions
\[\left\{ {\left| {{B_D}{A_D}} \right\rangle ,\left| {{B_D}{A_U}} \right\rangle ,\left| {{B_D}{A_C}} \right\rangle ,\left| {{B_C}{A_D}} \right\rangle ,\left| {{B_C}{A_U}} \right\rangle ,\left| {{B_C}{A_C}} \right\rangle } \right\}\]
where, for example, ${\left| {{B_C}{A_D}} \right\rangle }$ symbolizes the event in which the player believes the opponent will cooperate but the player intends to act by defecting. The model assumes that at the beginning of a game, the person has some potential to be in every circle in Fig. \ref{sixstate} which illustrates the possible transitions among the six states. So the person's state is a superposition of the six basis states
\begin{equation}
\begin{array}{l}
\left| \psi  \right\rangle  = {\psi _{DD}} \cdot \left| {{B_D}{A_D}} \right\rangle  + {\psi _{UD}} \cdot \left| {{B_D}{A_U}} \right\rangle  + {\psi _{CD}} \cdot \left| {{B_D}{A_C}} \right\rangle  + {\psi _{DC}} \cdot \left| {{B_C}{A_D}} \right\rangle \\
{\kern 1pt} {\kern 1pt} {\kern 1pt} {\kern 1pt} {\kern 1pt} {\kern 1pt} {\kern 1pt} {\kern 1pt} {\kern 1pt} {\kern 1pt} {\kern 1pt} {\kern 1pt} {\kern 1pt} {\kern 1pt} {\kern 1pt} {\kern 1pt} {\kern 1pt} {\kern 1pt}  + {\psi _{UC}} \cdot \left| {{B_C}{A_U}} \right\rangle  + {\psi _{CC}} \cdot \left| {{B_C}{A_C}} \right\rangle
\end{array}
\end{equation}
and the initial state corresponds to an amplitude distribution represented by the $6 \times 1$ column matrix
\[\psi \left( {\rm{0}} \right){\rm{ = }}\left[ {\begin{array}{*{20}{c}}
{{\psi _{DD}}}\\
{{\psi _{UD}}}\\
{{\psi _{CD}}}\\
{{\psi _{DC}}}\\
{{\psi _{UC}}}\\
{{\psi _{CC}}}
\end{array}} \right].\]
where, for example, ${{{\left| {{\psi _{DC}}} \right|}^2}}$ is the probability of observing state $\left| {{B_C}{A_D}} \right\rangle $ initially. The squared length of $\phi$ must equal one: ${\psi ^\dag } \cdot \psi  = 1$.

\textbf{Step 3: Inferences based on prior information }

During the decision process, new information at time ${t_1}$ changes the initial state ar time $t=0$ into a new state at time ${t_1}$.
For example, if the opponent is known to defect, the amplitude distribution across states changes to
\begin{equation}
\psi \left( {{t_1}} \right) = \frac{1}{{\sqrt {{{\left| {{\psi _{DD}}} \right|}^2} + {{\left| {{\psi _{UD}}} \right|}^2} + {{\left| {{\psi _{CD}}} \right|}^2}} }}\left[ {\begin{array}{*{20}{c}}
{{\psi _{DD}}}\\
{{\psi _{UD}}}\\
{{\psi _{CD}}}\\
0\\
0\\
0
\end{array}} \right] = \left[ {\begin{array}{*{20}{c}}
{{\psi _D}}\\
\textbf{0}
\end{array}} \right].
\end{equation}
where ${{{\left| {{\psi _{DD}}} \right|}^2} + {{\left| {{\psi _{UD}}} \right|}^2} + {{\left| {{\psi _{CD}}} \right|}^2}}$ equals the initial probability that the opponent defects (before given any information).

However, if the opponent's information is totally unknown, the amplitude distribution will remain the same as the initial $t=0$.
\[\begin{array}{l}
\psi \left( {{t_1}} \right) = \psi \left( 0 \right) = \left[ {\begin{array}{*{20}{c}}
{\sqrt {{{\left| {{\psi _{DD}}} \right|}^2} + {{\left| {{\psi _{UD}}} \right|}^2} + {{\left| {{\psi _{CD}}} \right|}^2}}  \cdot {\psi _D}}\\
{\sqrt {{{\left| {{\psi _{DC}}} \right|}^2} + {{\left| {{\psi _{UC}}} \right|}^2} + {{\left| {{\psi _{CC}}} \right|}^2}}  \cdot {\psi _C}}
\end{array}} \right]\\
= \sqrt {{{\left| {{\psi _{DD}}} \right|}^2} + {{\left| {{\psi _{UD}}} \right|}^2} + {{\left| {{\psi _{CD}}} \right|}^2}}  \cdot \left[ {\begin{array}{*{20}{c}}
{{\psi _D}}\\
\textbf{0}
\end{array}} \right] + \sqrt {{{\left| {{\psi _{DC}}} \right|}^2} + {{\left| {{\psi _{UC}}} \right|}^2} + {{\left| {{\psi _{CC}}} \right|}^2}}  \cdot \left[ {\begin{array}{*{20}{c}}
\textbf{0}\\
{{\psi _C}}
\end{array}} \right]
\end{array}\]
The equation shows that the initial state is a superposition formed by a weighted sum of the amplitude distribution for the two known conditions.

\textbf{Step 4: Obtain BPA based on payoffs}

During the decision making process, the participants need to evaluate the payoffs in order to select an appropriate action, which evolve the state at time ${t_1}$ into a new state at time ${t_2}$. The evolution of the state during this time period corresponds the thought process leading to a action decision, defection, uncertainty or cooperation.
Based on the preliminaries, an unitary matrix $U = {e^{ - iHt}}$ is defined to satisfy the solution of Schr${\ddot o}$dinger equation.
\[\psi \left( {{t_2}} \right) = {e^{ - iHt}} \cdot \psi \left( {{t_1}} \right)\]
$\psi \left( {{t_2}} \right)$ is the amplitude distribution across states after evolution based on payoffs.
\begin{equation}
H = \left[ {\begin{array}{*{20}{c}}
{{H_D}}&0\\
0&{{H_C}}
\end{array}} \right]
\end{equation}
with
\begin{equation}
{H_D} = \left( {\begin{array}{*{20}{c}}
{{h_D}}&0&1\\
0&1&0\\
1&0&{ - {h_D}}
\end{array}} \right)
\end{equation}
and
\begin{equation}
{H_C} = \left( {\begin{array}{*{20}{c}}
{{h_C}}&0&1\\
0&1&0\\
1&0&{ - {h_C}}
\end{array}} \right)
\end{equation}
The $3 \times 3$ Hamiltonian matrix ${H_D}$ applies when the participant believes the opponent will defect, and the other $3 \times 3$ Hamiltonian matrix ${H_C}$ applies when the participant believes the opponent will cooperate. The parameter ${h_D}$ is a function of the difference between the reward for defecting relative to cooperating given the opponent will defect, and ${h_C}$ is a function of the difference between the reward for defecting relative to cooperating given the opponent will cooperate. The Hamiltonian matrix $H$ transforms the state probabilities to favor defection, cooperation or uncertainty, depending on the reward function.

Each decision corresponds to a measurement of the state at time ${t_2}$. To obtain the BPA of certain decision, we can use an according measurement matrix $M$ to pick out the state. As the state $\psi$ is a $6 \times 1$ column matrix, the measurement matrix is defined as a $6 \times 6$ one.
\begin{equation}
M = \left( {\begin{array}{*{20}{c}}
{{M_D}}&\textbf{0}\\
\textbf{0}&{{M_C}}
\end{array}} \right),
\end{equation}
where all the elements is in the form of a $3 \times 3$ matrix. ${{M_D}}$ is used for measuring when believing the opponent will defect, and ${{M_C}}$ is used for measuring when believing the opponent will cooperate.
For example, to obtain the BPA of defecting, the measurement matrix is defined as
\[{M_D} = {M_C} = \left[ {\begin{array}{*{20}{c}}
1&0&0\\
0&0&0\\
0&0&0
\end{array}} \right]\]
Then we can obtain the BPA of defecting in unknown condition using Eq.(\ref{measure defect}).
\begin{equation}\label{measure defect}
m\left( DU \right) = M \cdot \psi \left( {{t_2}} \right)
\end{equation}
The BPA of defecting in the condition that the opponent is known to defect can be obtained by Eq.(\ref{measure defect2}).
\begin{equation}\label{measure defect2}
m\left( DD \right) = ({{{\left| {{\psi _{DD}}} \right|}^2} + {{\left| {{\psi _{UD}}} \right|}^2} + {{\left| {{\psi _{CD}}} \right|}^2}}) \cdot M \cdot \psi \left( {{t_2}} \right)
\end{equation}

\textbf{Step 5: Calculate entanglement degree}

Now, the BPAs of decisions have been obtained.
The BPA in known condition is
\begin{equation}
{m_1} = \left\{ {{m_1}\left( {DD} \right),{m_1}\left( {UD} \right),{m_1}(CD),{m_1}\left( {DC} \right),{m_1}\left( {UC} \right),{m_1}(CC)} \right\}.
\end{equation}
As the participant's beliefs are certain, no interference effect will be produced. The BPA of uncertain state is distributed based on classical $PPT$. The cardinality of $DD$, $CD$, $DC$ and $CC$ is 1. The cardinality of $UD$ and $UC$ is 2 as the actions are uncertain.

The BPA in unknown condition is
\begin{equation}
{m_2} = \left\{ {{m_2}\left( DU \right),{m_2}\left( UU \right),{m_2}(CU)} \right\}.
\end{equation}
It should be noticed that
\[{m_2}\left( DU \right) = {m_1}\left( {DD} \right) + {m_1}\left( {DC} \right)\]
\[{m_2}\left( UU \right) = {m_1}\left( {UD} \right) + {m_1}\left( {UC} \right)\]
\[{m_2}\left( CU \right) = {m_1}\left( {CD} \right) + {m_1}\left( {CC} \right)\]
As the opponent's decision is totally unknown and entanglement exists between beliefs and action, the BPA of uncertain state is distributed based on an entanglement degree. Then the interference effect will be produced. The cardinality of $D$ and $C$ is 2 as the beliefs are uncertain. The cardinality of $UU$ is 3 as both the beliefs and actions are uncertain.

Then an entanglement degree is defined in our model using Deng entropy.
\begin{definition}
Let ${m_1}$ and ${m_2}$ be two BPAs which illustrate the supporting of decisions under different frames of discernment. ${m_1}$ is under the known condition and ${m_2}$ is under the unknown condition. ${E_{d1}}$ and ${E_{d2}}$ are information volume of ${m_1}$ and ${m_2}$ respectively measured by Deng entropy. The entanglement degree $\gamma $ is defined as follows:
\begin{equation}\label{entanglement rate}
\gamma  = \frac{{{E_{d2}} - {E_{d1}}}}{{{E_{d2}}}}
\end{equation}
\end{definition}

%Before calculating the entanglement rate of a BPA, let us take a example of students taking course to illustrate the link between entanglement and Deng entropy. Suppose there are 32 students participating in a course examination. Each student has the same probability of winning the best. To find the best student, we are allowed to ask the teacher questions and the teacher will only answer yes or no. The problem is that how many times need at most to find the best student. Assume the times is t, it is easy to answer the problem through calculating the information volume by using information entropy

\textbf{Step 6: Obtain probability distribution}

As the uncertain state is not an outcome of the game, the BPA of uncertain state should be distributed into cooperation state and defection state.
For BPA ${m_1}$,
\begin{equation}\label{PPTdefect}
{P_1}\left( D \right) = {m_1}\left( {DD} \right) + \frac{1}{2}{m_1}\left( {UD} \right) + {m_1}\left( {DC} \right) + \frac{1}{2}{m_1}\left( {UC} \right).
\end{equation}
For BPA ${m_2}$,
\begin{equation}\label{entangle defect}
{P_2}\left( D \right) = {m_2}\left( DU \right) + \left( {\frac{1}{2} \pm \gamma } \right){m_2}\left( UU \right).
\end{equation}
where $\gamma $ is an entanglement degree, $ \pm $ corresponds to produce positive interference effect and negative interference effect respectively.

Obviously, ${P_1}\left( D \right)$ is different from ${P_2}\left( D \right)$. The difference value is the interference effect $Int$.
\begin{equation}
Int = {P_2}\left( D \right) - {P_1}\left( D \right) =  \pm \gamma  \cdot {m_2}\left( UU \right)
\end{equation}
In sum, based on the above steps, the decision making process is modelled and the interference effect can be effectively measured.
\section{Categorization decision-making experiment}
\subsection{Experiment}
Townsend $et al.$\cite{Townsend2000Exploring} proposed a categorization decision-making experiment to study the interactions between categorization and decision making. In the experiment, pictures of faces varying along face width and lip thickness are shown to participants. Generally, the faces can be distributed into two different kinds: one is a narrow (N) face with a narrow width and thick lips; the other one is a wide (W) face with a wide width and thin lips (see Figure \ref{face} for example). The participants are informed that N face had a 0.60 probability to come from the "bad" guy population while W face had a 0.60 probability to come from the "good" guy population.
\begin{figure}[!ht]
\centering
\includegraphics[scale=1]{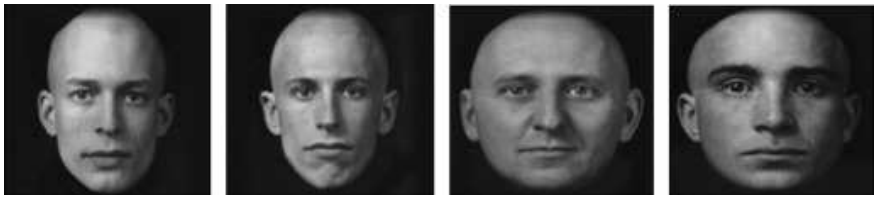}
 \caption{Example faces used in a categorization decision-making experiment}\label{face}
\end{figure}
The experiment can be classified into two parts. One part is in categorization-decision making (C-D) condition: participants are asked to categorize a face as belonging to either a "good" (G) guy or "bad" (B) guy following make a decision whether to "attack" (A) or to to "withdraw" (W). The other part is decision-making along (D-along) condition: participants are asked to only make an action decision.
The experiment included a total of 26 participants, but each participant provided 51 observations for the C-D condition for a total of ${\rm{26}} \times {\rm{51 = 1326}}$ observations, while each person produced 17 observations for the D condition for a total of ${\rm{17}} \times {\rm{26 = 442}}$ total observations.

The experiment results are shown in Table \ref{classical result}.
\begin{table}[!h]
\centering
\caption{The results of C-D condition and D-along condition}
\label{classical result}
\begin{tabular}{cccccccc}
\toprule
Type face & \textbf{$P\left( {G} \right)$} & \textbf{$P\left( {A|G} \right)$} & \textbf{$P\left( {B} \right)$} & \textbf{$P\left( {A|B} \right)$} & \textbf{${P_T}$} & \textbf{$P\left( {A} \right)$}  \\
\midrule
Wide               & 0.84          & 0.35            & 0.16          & 0.52            & 0.37       & 0.39          \\
Narrow             & 0.17          & 0.41            & 0.83          & 0.63            & 0.59       & 0.69          \\
\bottomrule
\end{tabular}
\end{table}
The column labeled $P\left( {G} \right)$ represents the probability of categorizing the face as a "good buy".
The column labeled $P\left( {A|G} \right)$ represents the probability of attacking given categorizing the face as a "good guy".
The column labeled $P\left( {B} \right)$ represents the probability of categorizing the face as a "bad buy".
The column labeled $P\left( {A|B} \right)$ represents the probability of attacking given categorizing the face as a "bad guy".
And the column labeled ${P_T}$ represents the final probability of attacking in C-D condition which is computed as follows:
\begin{equation}\label{Pt}
P\left( A \right) = P(G) \cdot P\left( {A|G} \right) + P\left( B \right) \cdot P\left( {A|B} \right)
\end{equation}
Accordingly, the column label as $P\left( {A} \right)$ represents the probability of attacking in D-along condition. As shown in Table \ref{classical result}, some deviation exist between ${P_T}$ and $P\left( {A} \right)$ for both types of face which is called the disjunction effect. However, the disjunction effect is prominent for narrow type faces, while the disjunction effect is weak for wide type faces.

The classical paradigm has been discussed in many works. Literatures of studying the categorization decision-making experiment and their results are shown below in Table \ref{preworks}.
\begin{table}[!h]
\centering
\small
\caption{Results of other categorization decision-making experiments}
\label{preworks}
\begin{threeparttable}
\begin{tabular*}{\columnwidth}{ccccccccc}
\toprule
Literature                                                                                           & Type & $P\left( {G} \right)$ & $P\left( {A|G} \right)$ & $P\left( {B} \right)$ & $P\left( {A|B} \right)$ & ${P_T}$   & $P\left( {A} \right)$   \\
\midrule
\multirow{2}{*}{\scriptsize\begin{tabular}[c]{@{}c@{}}Townsend $et al.$(2000)\cite{Townsend2000Exploring}\end{tabular}}                     & W    & 0.84 & 0.35   & 0.16 & 0.52   & 0.37 & 0.39  \\
                                                                                                     & N    & 0.17 & 0.41   & 0.83 & 0.63   & 0.59 & 0.69   \\ \hline
\multirow{2}{*}{\scriptsize\begin{tabular}[c]{@{}c@{}}Busemeyer $et al.$(2009)\cite{Busemeyer2009Empirical}\end{tabular}}                    & W    & 0.80 & 0.37   & 0.20 & 0.53   & 0.40 & 0.39  \\
                                                                                                     & N    & 0.20 & 0.45   & 0.80 & 0.64   & 0.60 & 0.69   \\ \hline
\multirow{2}{*}{\scriptsize\begin{tabular}[c]{@{}c@{}}Wang and Busemeyer(2016)\\  Experiment 1\cite{wang2016interference}\end{tabular}} & W    & 0.78 & 0.39   & 0.22 & 0.52   & 0.42 & 0.42      \\
                                                                                                     & N    & 0.21 & 0.41   & 0.79 & 0.58   & 0.54 & 0.59   \\ \hline
\multirow{2}{*}{\scriptsize\begin{tabular}[c]{@{}c@{}}Wang and Busemeyer(2016) \\ Experiment 2\cite{wang2016interference}\end{tabular}} & W    & 0.78 & 0.33   & 0.22 & 0.53   & 0.37 & 0.37     \\
                                                                                                     & N    & 0.24 & 0.37   & 0.76 & 0.61   & 0.55 & 0.60   \\ \hline
\multirow{2}{*}{\scriptsize\begin{tabular}[c]{@{}c@{}}Wang and Busemeyer(2016) \\ Experiment 3\cite{wang2016interference}\end{tabular}} & W    & 0.77 & 0.34   & 0.23 & 0.58   & 0.40 & 0.39  \\
                                                                                                     & N    & 0.24 & 0.33   & 0.76 & 0.66   & 0.58 & 0.62   \\\hline
\multirow{2}{*}{Average}                                                                             & W    & 0.79 & 0.36   & 0.21 & 0.54   & 0.39 & 0.39      \\
                                                                                                     & N    & 0.21 & 0.39   & 0.79 & 0.62   & 0.57 & 0.64 \\
\bottomrule
\end{tabular*}
 \begin{tablenotes}
        \footnotesize
        \item[1] In Busemeyer $et al.$(2009)\cite{Busemeyer2009Empirical}, the classical experiment is replicated.
        \item[2] In Wang and Busemeyer(2016)\cite{wang2016interference}, Experiment 1 uses a larger data set to replicate the classical experiment. Experiment 2 introduce a new X-D trial verse C-D trial and only the result of C-D trial is used here. In experiment 3, the reward for attacking bad people is a bit less than the other two.
      \end{tablenotes}
\end{threeparttable}
\end{table}

\subsection{Application}
In the following, our model will be applied to explain the disjunction effect for narrow type faces.

\textbf{Step 1: Determine the frame of discrimination}

In C-D condition, the outcome will be either to A or W given categorizing the face as G or B. Thus the basic states "AG","WG","AB" and "WB" consist the frame of discernment, where, for example, "AG" represents the participant decide to attack when categorizing the face as a good guy. Then extra state "AWG", "AWB" and $\emptyset$ will be included to fill the power set of this discrimination. State "AWG" represents that the participant is uncertain to attack or withdraw given categorizing the face as a good guy and it can be denoted as "UG". Same as it, state "AWB" can be denoted as "UB". And the empty set $\emptyset$ is irrespective in the model as it is meaningless.

In D-along condition, the outcome of the game will be either to A or W without categorizing the face. Thus the basic states "AU" and "WU" consist the frame of discrimination.
Then extra state "AWU" and $\emptyset$ will be included to fill the power set of this discrimination. State "AWU" presents that the participant is uncertain to attack or withdraw without categorization and it can be denoted as "UU". And the empty set $\emptyset$ is irrespective in the model as it is meaningless.

\textbf{Step 2: Representation of beliefs and actions}

The initial state involves six combination of beliefs and actions
\[\left\{ {\left| {{B_G}{A_A}} \right\rangle ,\left| {{B_G}{A_U}} \right\rangle ,\left| {{B_G}{A_W}} \right\rangle ,\left| {{B_B}{A_A}} \right\rangle ,\left| {{B_B}{A_U}} \right\rangle ,\left| {{B_C}{A_W}} \right\rangle } \right\}\]. Participants' state is a superposition of the six basis states
\begin{equation}
\begin{array}{l}
\left| \psi  \right\rangle  = {\psi _{DD}} \cdot \left| {{B_D}{A_D}} \right\rangle  + {\psi _{UD}} \cdot \left| {{B_D}{A_U}} \right\rangle  + {\psi _{CD}} \cdot \left| {{B_D}{A_C}} \right\rangle  + {\psi _{DC}} \cdot \left| {{B_C}{A_D}} \right\rangle \\
{\kern 1pt} {\kern 1pt} {\kern 1pt} {\kern 1pt} {\kern 1pt} {\kern 1pt} {\kern 1pt} {\kern 1pt} {\kern 1pt} {\kern 1pt} {\kern 1pt} {\kern 1pt} {\kern 1pt} {\kern 1pt} {\kern 1pt} {\kern 1pt} {\kern 1pt} {\kern 1pt}  + {\psi _{UC}} \cdot \left| {{B_C}{A_U}} \right\rangle  + {\psi _{CC}} \cdot \left| {{B_C}{A_D}} \right\rangle
\end{array}
\end{equation}
and the initial state corresponds to an amplitude distribution represented by the $6 \times {\rm{1}}$ column matrix
\[\psi \left( {\rm{0}} \right){\rm{ = }}\left[ {\begin{array}{*{20}{c}}
{{\psi _{AG}}}\\
{{\psi _{UG}}}\\
{{\psi _{WG}}}\\
{{\psi _{AB}}}\\
{{\psi _{UB}}}\\
{{\psi _{WB}}}
\end{array}} \right].\]

\textbf{Step 3: Inferences based on prior information}

In C-D condition, if the participant categorizes the face as "good", the state changes to
\begin{equation}
\psi \left( {{t_1}} \right) = \frac{1}{{\sqrt {{{\left| {{\psi _{AG}}} \right|}^2} + {{\left| {{\psi _{UG}}} \right|}^2} + {{\left| {{\psi _{WG}}} \right|}^2}} }}\left[ {\begin{array}{*{20}{c}}
{{\psi _{AG}}}\\
{{\psi _{UG}}}\\
{{\psi _{WG}}}\\
0\\
0\\
0
\end{array}} \right] = \left[ {\begin{array}{*{20}{c}}
{{\psi _G}}\\
\textbf{0}
\end{array}} \right].
\end{equation}
If the participant categorizes the face as "bad", the state changes to
\begin{equation}
\psi \left( {{t_1}} \right) = \frac{1}{{\sqrt {{{\left| {{\psi _{AB}}} \right|}^2} + {{\left| {{\psi _{UB}}} \right|}^2} + {{\left| {{\psi _{WB}}} \right|}^2}} }}\left[ {\begin{array}{*{20}{c}}
0\\
0\\
0\\
{{\psi _{AB}}}\\
{{\psi _{UB}}}\\
{{\psi _{WB}}}
\end{array}} \right] = \left[ {\begin{array}{*{20}{c}}
\textbf{0}\\
{{\psi _B}}
\end{array}} \right].
\end{equation}

In D-along condition, the state is a superposition formed by a weighted sum of the amplitude distribution for two known conditions in C-D condition.
\begin{equation}
\begin{array}{l}
\psi \left( {{t_1}} \right) = \sqrt {{{\left| {{\psi _{AG}}} \right|}^2} + {{\left| {{\psi _{UG}}} \right|}^2} + {{\left| {{\psi _{WG}}} \right|}^2}} \left[ {\begin{array}{*{20}{c}}
{{\psi _G}}\\
\textbf{0}
\end{array}} \right]\\
{\kern 29pt}  + \sqrt {{{\left| {{\psi _{AB}}} \right|}^2} + {{\left| {{\psi _{UB}}} \right|}^2} + {{\left| {{\psi _{WB}}} \right|}^2}} \left[ {\begin{array}{*{20}{c}}
\textbf{0}\\
{{\psi _B}}
\end{array}} \right].
\end{array}
\end{equation}

\textbf{Step 4: Obtain BPA based on payoffs}

The evolution of the state obeys a Schr${\ddot o}$dinger equation (Eq. (\ref{schrodinger})). The solution is
\[\psi \left( {{t_2}} \right) = {e^{ - iHt}} \cdot \psi \left( {{t_1}} \right)\]
In this problem, $t$ is set to $\frac{\pi }{2}$ as Busemeyer $et al.$\cite{Busemeyer2009Empirical}. The Hamiltonian matrix $H$ is
\begin{equation}
H = \left[ {\begin{array}{*{20}{c}}
{{H_G}}&0\\
0&{{H_B}}
\end{array}} \right]
\end{equation}
with
\[{H_G} = \left( {\begin{array}{*{20}{c}}
{{h_G}}&0&1\\
0&1&0\\
1&0&{ - {h_G}}
\end{array}} \right){\kern 1pt} {\kern 1pt} {\kern 1pt} {\kern 1pt} and{\kern 1pt} {\kern 1pt} {\kern 1pt} {H_B} = \left( {\begin{array}{*{20}{c}}
{{h_B}}&0&1\\
0&1&0\\
1&0&{ - {h_B}}
\end{array}} \right)\]
%\begin{equation}
%{H_G} = \left( {\begin{array}{*{20}{c}}
%{{h_G}}&0&1\\
%0&1&0\\
%1&0&{ - {h_G}}
%\end{array}} \right)
%\end{equation}
%and
%\begin{equation}
%{H_B} = \left( {\begin{array}{*{20}{c}}
%{{h_B}}&0&1\\
%0&1&0\\
%1&0&{ - {h_B}}
%\end{array}} \right)
%\end{equation}

In C-D condition, if the face is categorized as "good", the state changes to
\begin{equation}
\psi \left( {{t_2}} \right) = {e^{ - i \cdot H \cdot t}} \cdot \psi ({t_{\rm{1}}}){\rm{ = }}\left[ {\begin{array}{*{20}{c}}
{{e^{ - i \cdot {H_G} \cdot t}}}&0\\
0&{{e^{ - i \cdot {H_B} \cdot t}}}
\end{array}} \right] \cdot \left[ {\begin{array}{*{20}{c}}
{{\psi _G}}\\
\textbf{0}
\end{array}} \right] = {e^{ - i \cdot {H_G} \cdot t}} \cdot {\psi _G}
\end{equation}
If the face is categorized as "bad", the state changes to
\begin{equation}
\psi \left( {{t_2}} \right) = {e^{ - i \cdot H \cdot t}} \cdot \psi ({t_{\rm{1}}}){\rm{ = }}\left[ {\begin{array}{*{20}{c}}
{{e^{ - i \cdot {H_G} \cdot t}}}&0\\
0&{{e^{ - i \cdot {H_B} \cdot t}}}
\end{array}} \right] \cdot \left[ {\begin{array}{*{20}{c}}
\textbf{0}\\
{{\psi _B}}
\end{array}} \right] = {e^{ - i \cdot {H_B} \cdot t}} \cdot {\psi _B}
\end{equation}
In D-along condition, the state changes to
\begin{equation}
\footnotesize
\begin{array}{*{20}{l}}
{\psi \left( {{t_2}} \right) = {e^{ - i \cdot H \cdot t}} \cdot \psi (0){\rm{ = }}\left[ {\begin{array}{*{20}{c}}
{{e^{ - i \cdot {H_G} \cdot t}}}&0\\
0&{{e^{ - i \cdot {H_B} \cdot t}}}
\end{array}} \right] \cdot \left[ {\begin{array}{*{20}{c}}
{\sqrt {{{\left| {{\psi _{AG}}} \right|}^2} + {{\left| {{\psi _{UG}}} \right|}^2} + {{\left| {{\psi _{WG}}} \right|}^2}}  \cdot {\psi _G}}\\
{\sqrt {{{\left| {{\psi _{AB}}} \right|}^2} + {{\left| {{\psi _{UB}}} \right|}^2} + {{\left| {{\psi _{WB}}} \right|}^2}}  \cdot {\psi _B}}
\end{array}} \right]}\\
\begin{array}{l}
{\kern 1pt} {\kern 1pt} {\kern 1pt} {\kern 1pt} {\kern 1pt} {\kern 1pt} {\kern 1pt} {\kern 1pt} {\kern 1pt} {\kern 1pt} {\kern 1pt} {\kern 1pt} {\kern 1pt} {\kern 1pt} {\kern 1pt} {\kern 1pt} {\kern 1pt} {\kern 1pt} {\kern 1pt} {\kern 1pt} {\kern 1pt} {\kern 1pt} {\kern 1pt} {\kern 1pt} = \sqrt {{{\left| {{\psi _{AG}}} \right|}^2} + {{\left| {{\psi _{UG}}} \right|}^2} + {{\left| {{\psi _{WG}}} \right|}^2} \cdot } {e^{ - i \cdot {H_G} \cdot t}} \cdot {\psi _G}{\rm{ + }}\\
{\kern 1pt} {\kern 1pt} {\kern 1pt} {\kern 1pt} {\kern 1pt} {\kern 1pt} {\kern 1pt} {\kern 1pt} {\kern 1pt} {\kern 1pt} {\kern 1pt} {\kern 1pt} {\kern 1pt} {\kern 1pt} {\kern 1pt} {\kern 1pt} {\kern 1pt} {\kern 1pt} {\kern 1pt} {\kern 1pt} {\kern 1pt} {\kern 1pt} {\kern 1pt} {\kern 1pt} {\kern 1pt} {\kern 1pt} {\kern 1pt} {\kern 1pt} {\kern 1pt} {\kern 1pt} {\kern 1pt} {\kern 1pt} {\kern 1pt} {\kern 1pt} {\kern 1pt} {\kern 1pt} {\kern 1pt} {\kern 1pt} {\kern 1pt} {\kern 1pt} {\kern 1pt} \sqrt {{{\left| {{\psi _{AB}}} \right|}^2} + {{\left| {{\psi _{UB}}} \right|}^2} + {{\left| {{\psi _{WB}}} \right|}^2}}  \cdot {e^{ - i \cdot {H_B} \cdot t}} \cdot {\psi _B}
\end{array}
\end{array}
\end{equation}

The measure matrix is
\[M = \left( {\begin{array}{*{20}{c}}
{{M_1}}&0\\
0&{{M_2}}
\end{array}} \right)\]
where ${M_1}$ and ${M_2}$ is set as
\[{M_1} = {M_2} = \left[ {\begin{array}{*{20}{c}}
1&0&0\\
0&0&0\\
0&0&0
\end{array}} \right],\left[ {\begin{array}{*{20}{c}}
0&0&0\\
0&1&0\\
0&0&0
\end{array}} \right]or\left[ {\begin{array}{*{20}{c}}
0&0&0\\
0&0&0\\
0&0&1
\end{array}} \right]\]
respectively to pick out the state of attacking, uncertainty or defecting.

In C-D condition, the BPA of states "AG", "UG" and "WG" can be obtained by Eq. (\ref{P(G)*M}).
\begin{equation}\label{P(G)*M}
m = P\left( G \right) \cdot {\left\| {M \cdot {e^{ - itH}} \cdot \psi \left( {{t_1}} \right)} \right\|^2}
\end{equation}
where $P\left( G \right)$ is the probability that a face is categorized as "G", equalling to ${\sqrt {{{\left| {{\psi _{AG}}} \right|}^2} + {{\left| {{\psi _{UG}}} \right|}^2} + {{\left| {{\psi _{WG}}} \right|}^2}} }$.

The BPA of states "AB", "UB" and "WB" can be obtained by Eq. (\ref{P(B)*M}).
\begin{equation}\label{P(B)*M}
m = P\left( B \right) \cdot {\left\| {M \cdot {e^{ - itH}} \cdot \psi \left( {{t_1}} \right)} \right\|^2}
\end{equation}
where $P\left( B \right)$ is the probability that a face is categorized as "B", equalling to ${\sqrt {{{\left| {{\psi _{AB}}} \right|}^2} + {{\left| {{\psi _{UB}}} \right|}^2} + {{\left| {{\psi _{WB}}} \right|}^2}} }$.

Thus the BPA in C-D condition is
\[\begin{array}{l}
{m_1} = \left( {m\left( {AG} \right),m\left( {UG} \right),m\left( {WG} \right),m\left( {AB} \right),m\left( {UB} \right),m\left( {WG} \right)} \right)\\
 = \left( {0.0414,0.0567,0.0720,0.3846,0.2767,0.1688} \right).
\end{array}\]

In D-along condition, the BPA of states "AU", "UU" and "WU" can be obtained by Eq. (\ref{1*M}).
\begin{equation}\label{1*M}
m = {\left\| {M \cdot {e^{ - itH}} \cdot \psi \left( {{t_1}} \right)} \right\|^2}
\end{equation}
Thus the BPA in D-along condition is
\[{m_2} = \left( {m\left( AU \right),m\left( UU \right),m\left( WU \right)} \right) = \left( {0.4259,0.3333,0.2407} \right).\]

There are four free parameters existing in our model shown as Table \ref{parametervalue}. ${h_G}$ and ${h_B}$ are reward functions set by users. $P(G)$ and $P(B)$ are obtained from the experiment result.
\begin{table}[!h]
\centering
\caption{The value of free parameters}
\label{parametervalue}
\begin{tabular}{ccccc}
\toprule
Parameters & ${h_G}$ & ${h_B}$ & $P(G)$ & $P(B)$\\
\midrule
Value      & -0.1376  & 0.2033  & 0.83  & 0.17  \\
\bottomrule
\end{tabular}
\end{table}

\textbf{Step 5: Calculate entanglement degree}

The information volume of ${m_1}$ and ${m_2}$ is measured as following:
\begin{equation}
\begin{array}{l}
{E_{d1}} =  - 0.0414{\log _2}\left( {\frac{{0.0414}}{{{2^1} - 1}}} \right) - 0.0567{\log _2}\left( {\frac{{0.0567}}{{{2^2} - 1}}} \right) - 0.072{\log _2}\left( {\frac{{0.072}}{{{2^1} - 1}}} \right)\\
 - 0.3846{\log _2}\left( {\frac{{0.3846}}{{{2^1} - 1}}} \right) - 0.2767{\log _2}\left( {\frac{{0.2767}}{{{2^2} - 1}}} \right) - 0.1688{\log _2}\left( {\frac{{0.1688}}{{{2^1} - 1}}} \right)\\
{\kern 18pt} = 2.7026
\end{array}
\end{equation}
\begin{equation}
\begin{array}{l}
{E_{d2}} =  - 0.4259{\log _2}\left( {\frac{{0.4259}}{{{2^2} - 1}}} \right) - 0.3333{\log _2}\left( {\frac{{0.3333}}{{{2^3} - 1}}} \right) - 0.2407{\log _2}\left( {\frac{{0.2407}}{{{2^2} - 1}}} \right)\\
{\kern 18pt} = 3.5398
\end{array}
\end{equation}
Then then entanglement degree $\gamma $ is calculated as
\begin{equation}
\gamma  = \frac{{{E_{d2}} - {E_{d1}}}}{{{E_{d2}}}} = 0.2365.
\end{equation}

\textbf{Step 6: Obtain probability distribution}

In C-D condition, the probability of attack ${P_1}$ is
\[{P_1}\left( A \right) = {m_1}\left( {AG} \right) + \frac{1}{2}{m_1}\left( {UG} \right) + {m_1}\left( {AB} \right) + \frac{1}{2}{m_1}\left( {UB} \right) = 0.5927\]
which is exactly the same as the experiment result.

In D-along condition, as a positive interference effect is produced, the probability of attack ${P_2}$ is
\[{P_2}\left( A \right) = {m_2}(AU) + \left( {\frac{1}{2} + \gamma } \right){m_2}\left( {UU} \right) = 0.6715\]
which is close to the experiment result.

The interference effect produced by categorization is measured as
\[Int = {P_2}\left( A \right) - {P_1}\left( A \right) =  \gamma  \cdot {m_2}\left( UU \right)=0.0788\]

\subsection{Comparison}

Compare the obtained model results with the observed experiment results (for N type face), the model results are close to the practical situation, which verifies the correctness and effectiveness of our model. As Table \ref{result compare} shows, the interference effect is predicted and the average error rate is less than 1\%. The decision making process is modelled and the disjunction effect is well explained in our model.
\begin{table}[!h]
\centering
\footnotesize
\caption{The result of QDB model}
\label{result compare}
\begin{threeparttable}
\begin{tabular}{cccccccc}
\toprule
Literature                                                                                        &     & $P\left( G \right)$ & $P\left( {A|G} \right)$\tnote{1} & $P\left( B \right)$ & $P\left( {A|B} \right)$\tnote{2} & ${{P}_T}$     & $P\left( A \right)$   \\
\midrule
\multirow{2}{*}{\scriptsize\begin{tabular}[c]{@{}c@{}}Townsend \\ $et al.$(2000)\end{tabular}}                  & Obs & 0.17 & 0.41   & 0.83 & 0.63   & 0.59   & 0.69   \\
                                                                                                               & QDB & 0.17 & 0.4100   & 0.83 & 0.6301 & 0.5926 & 0.6715 \\ \hline
\multirow{2}{*}{\scriptsize\begin{tabular}[c]{@{}c@{}}Busemeyer \\ $et al.$(2009)\end{tabular}}                 & Obs & 0.20 & 0.45   & 0.80 & 0.64   & 0.60   & 0.69   \\
                                                                                                               & QDB & 0.20  & 0.4470 & 0.80 & 0.6340 & 0.5965     & 0.6689 \\ \hline
\multirow{2}{*}{\scriptsize\begin{tabular}[c]{@{}c@{}}Wang and Busemeyer(2016)\\  Experiment 1\end{tabular}} & Obs & 0.21 & 0.41     & 0.79 & 0.58   & 0.54   & 0.59   \\
                                                                                                             & QDB & 0.21 & 0.4148   & 0.79 & 0.5916 & 0.5544 & 0.6241 \\
                                                                                                             \hline
\multirow{2}{*}{\scriptsize\begin{tabular}[c]{@{}c@{}}Wang and Busemeyer(2016) \\ Experiment 2\end{tabular}} & Obs & 0.24 & 0.37   & 0.76 & 0.61   & 0.55   & 0.60   \\
                                                                                                             & QDB & 0.24 & 0.3720 & 0.76 & 0.6162 & 0.5575 & 0.6247 \\
                                                                                                             \hline
\multirow{2}{*}{\scriptsize\begin{tabular}[c]{@{}c@{}}Wang and Busemeyer(2016) \\ Experiment 3\end{tabular}} & Obs & 0.24 & 0.33   & 0.76 & 0.66   & 0.58   & 0.62   \\
                                                                                                             & QDB & 0.24 & 0.3381 & 0.76 & 0.6454 & 0.5716 & 0.6417 \\
                                                                                                  \hline
\multirow{2}{*}{Average}                                                                          & Obs & 0.21 & 0.39   & 0.79 & 0.62   & 0.57   & 0.64   \\
                                                                                                  & QDB & 0.21 & 0.3964 & 0.79 & 0.6235 & 0.5758 & 0.6462\\
\bottomrule
\end{tabular}
 \begin{tablenotes}
        \footnotesize
        \item[1] Obs represents the observed experiment result.
        \item[2] QDB represents the results of quantum dynamic belief model.
      \end{tablenotes}
\end{threeparttable}
\end{table}

In the following, the comparison among our quantum dynamic belief model, Markov belief-action (BA) model and quantum belief-action entanglement (BAE) model will be made.

Markov BA model was proposed by Townsend $et al.$(2000)\cite{Townsend2000Exploring} to do the category-decision task. The model assumes that the categorization and decision-making are two parts in the chain, namely the categorization depends only on the face while the action depends only on the categorization.
\begin{figure}[!ht]
\centering
\label{markov}
\includegraphics[scale=0.53]{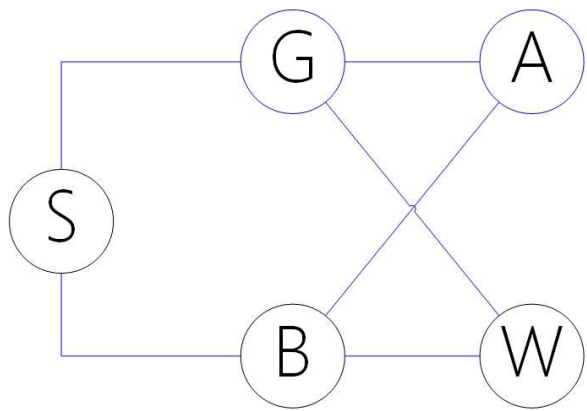}
\caption{The decision-making process of Markov BA model}\label{face}
\end{figure}
In C-D condition, the probability of attacking equals to $\phi (G) \cdot \phi (A|G)$ given the face is categorized as G and equals to $\phi (B) \cdot \phi (A|B)$ given the face is categorized as B.
In D-along condition, the probability of attacking equals to the probability of reaching the state $A$ by two different paths.
\[\phi (A) = \phi (G) \cdot \phi (A|G) + \phi (B) \cdot \phi (A|B)\]
Hence, the Markov BA model follows the law of total probability, which means that disjunction effect cannot be explained.

\begin{figure}[!ht]
\centering
\includegraphics[scale=0.7]{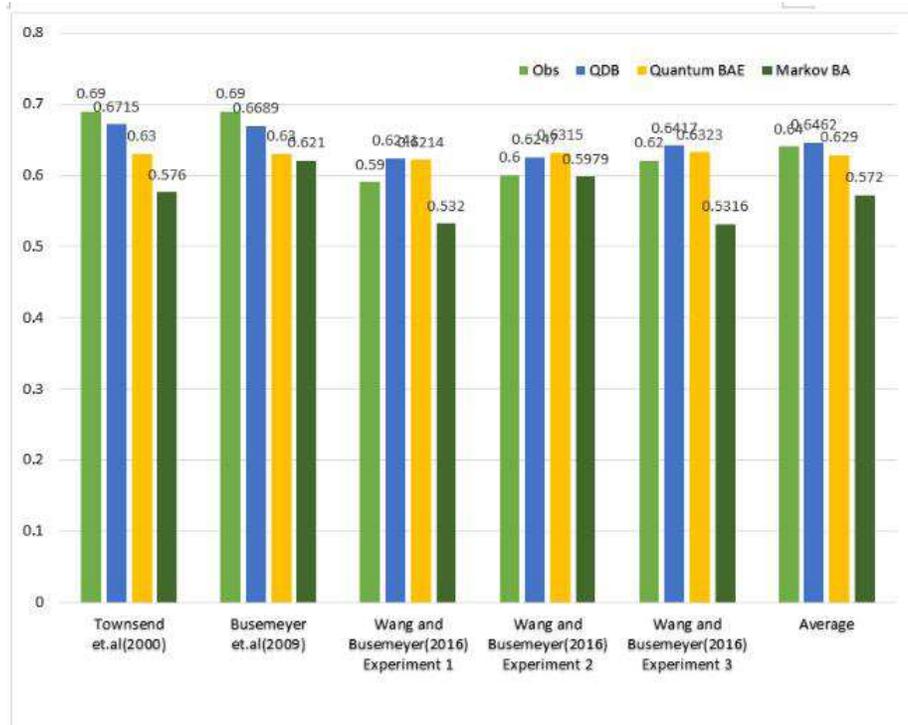}
 \caption{The comparison of probability of attacking in D-along condition}\label{zhuzhuangtu}
\end{figure}

Quantum BAE model was initially proposed by Pothos and Busemeyer (2009)\cite{Pothos2009A} and it was improved by Wang and Busemeyer (2016)\cite{wang2016interference}. The model bases on quantum dynamic modelling and the crucial factor to produce interference effect is that a free entanglement parameter $c$ is defined.
The unitary matrix in Eq. (\ref{psit_2}) is ${e^{ - i({H_1} + {H_2})t}}$, where ${H_1}$ is the same as $H$ in our model. ${H_2}$ rotates inferences for G face to match W action and rotates inferences for B face to match A action, which produces the interference effect.
\[{H_2} = \frac{c}{{\sqrt 2 }}\left[ {\begin{array}{*{20}{c}}
{ - 1}&0&1&0\\
0&1&0&1\\
1&0&1&0\\
0&1&0&{ - 1}
\end{array}} \right]\]
In order to compare the ability of predict the interference effect, we apply three models to obtain the probability of attacking in D-along condition. The comparison among the three models is shown in Fig. \ref{zhuzhuangtu}.
Both the quantum BAE model and our quantum dynamic belief model can predict the interference effect while the Markov BA model could not. However, the prediction result of our quantum dynamic belief model is more accurate. It should be noticed that although an extra uncertain state is introduced in our model, the number of free parameters decreases on the contrary. The entanglement degree is measured by an entropy function in our model rather a free parameter in the quantum BAE model. Based on the above, it is reasonable to conclude that the QDB model is correct and efficient.
\section{Conclusion}\label{Conclusion}
To explain the disjunction fallacy, a new quantum dynamic belief decision making model is proposed in this paper. The model combines Dempster-Shafer evidence theory with the quantum dynamic model. The uncertainty of belief is represented by a superposition of certain states. And the uncertainty in actions is represented as an extra uncertain state with D-S theory. The entanglement between beliefs and actions can produce the interference effect when the beliefs are uncertain. The interference effect is measured by the uncertain state in actions and an entanglement degree defined by Deng entropy. A classical categorization decision-making experiment is illustrated and the new model modelling a real human decision-making process can well explain the disjunction effect. In the end, the comparison with other models is made, which shows the correctness and effectiveness of our model.
\section{Acknowledgement}
The work is partially supported by National Natural Science Foundation of China (Grant No. 61671384), Natural Science Basic Research Plan in Shaanxi Province of China (Program No. 2016JM6018), Aviation Science Foundation (Program No. 20165553036), the Fund of SAST (Program No. SAST2016083)

\bibliographystyle{model1-num-names}
\bibliography{myreference}

\end{document}